\documentclass{article}

\usepackage{microtype}
\usepackage{graphicx}
\usepackage{subcaption}
\usepackage{booktabs} 
\usepackage{listings}
\usepackage{tabularx}
\usepackage{placeins}

\usepackage{xcolor}

\definecolor{codecomment}{RGB}{0,128,0}
\definecolor{codenumber}{RGB}{9,134,88}
\definecolor{codestring}{RGB}{163,21,21}
\definecolor{codekey}{RGB}{175,0,219}

\lstdefinestyle{mystyle}{
    commentstyle=\color{codecomment},
    keywordstyle=\color{codekey},
    numberstyle=\tiny\color{codenumber},
    stringstyle=\color{codestring},
    basicstyle=\ttfamily\scriptsize,
    breakatwhitespace=false,         
    breaklines=false,                 
    keepspaces=true,                 
    showspaces=false,                
    showstringspaces=false,
    showtabs=false,                  
    tabsize=2,
    columns=fullflexible,
}
 
\newcommand{\ic}[1]{\lstinline[basicstyle=\ttfamily\footnotesize]{#1}}

\usepackage{hyperref}

\usepackage[accepted]{mlsys2022}
\usepackage{inconsolata}

\begin{document}

\twocolumn[
\mlsystitle{torch.fx: Practical Program Capture and Transformation for Deep Learning in Python}



\mlsyssetsymbol{equal}{*}

\begin{mlsysauthorlist}
\mlsysauthor{James K Reed}{fb}
\mlsysauthor{Zachary DeVito}{fb}
\mlsysauthor{Horace He}{fb}
\mlsysauthor{Ansley Ussery}{fb}
\mlsysauthor{Jason Ansel}{fb}
\end{mlsysauthorlist}

\mlsysaffiliation{fb}{Facebook, Menlo Park, CA, USA}

\mlsyscorrespondingauthor{James K Reed}{jamesreed@fb.com}

\mlsyskeywords{Machine Learning, MLSys}

\vskip 0.3in

\begin{abstract}

Modern deep learning frameworks provide imperative, \emph{eager execution} programming interfaces embedded in Python to provide a productive development experience. However, deep learning practitioners sometimes need to capture and transform program structure for performance optimization, visualization, analysis, and hardware integration. We study the different designs for program capture and transformation used in deep learning. By designing for typical deep learning use cases rather than long tail ones, it is possible to create a simpler framework for program capture and transformation. We apply this principle in \ic{torch.fx}, a program capture and transformation library for PyTorch written entirely in Python and optimized for high developer productivity by ML practitioners. We present case studies showing how \ic{torch.fx} enables workflows previously inaccessible in the PyTorch ecosystem.

\end{abstract}
]



\printAffiliationsAndNotice{}  

\section{Introduction}
\label{introduction}

Early \emph{graph mode} or \emph{define-and-run}~\cite{DBLP:journals/corr/abs-1908-00213} deep learning frameworks like Caffe~\cite{DBLP:journals/corr/JiaSDKLGGD14}, Theano~\cite{DBLP:journals/corr/Al-RfouAAa16}, and TensorFlow~\cite{45381} defined APIs in which the user constructed a graph-based intermediate representation (IR) of the desired computation. Program transformations like program differentiation, device/host partitioning and placement, quantization, device lowering, and performance optimization could be applied directly to this IR. One way to think of these frameworks is as simple embedded programming languages that are meta-programmed from a host language, predominantly Python~\cite{innes_2017}.

However, these frameworks require the user to exit the host language and enter a domain-specific language and runtime, which often has inferior user experience compared to the host language. For instance, debugging requires different tools from the typical debugging toolkits such as Python's \ic{pdb} library.

More recent \textit{eager mode} or \textit{define-by-run}~\cite{DBLP:journals/corr/abs-1908-00213} frameworks such as Autograd~\cite{maclaurinautograd}, Chainer~\cite{DBLP:journals/corr/abs-1908-00213}, PyTorch~\cite{DBLP:journals/corr/abs-1912-01703} and TensorFlow Eager~\cite{DBLP:journals/corr/abs-1903-01855} eschew explicit graph-building APIs in favor of programming in the host language directly. The primary program transformation used in deep learning frameworks, program differentiation, is reformulated from an ahead-of-time transformation to a just-in-time transformation, in the form of auto-differentiation.

Most training and inference can be done using eager mode with auto-differentiation. However, there are still transformations\textemdash such as program quantization or operator fusion\textemdash that are easier to write given the additional program structure provided by an IR. To bridge this gap, an eager-mode framework needs a way of capturing program structure from user programs to enable these transformations.

Some program capture systems are built to capture a freestanding representation of the whole program for the purposes of serialization or export. For instance, TorchScript~\cite{torchscript} includes mutable state, control-flow, and complex data types for the purposes of faithfully modeling the semantics of the original Python program. Modeling Python in full generality comes at the cost of complexity in program capture techniques and difficulty of writing transforms on the highly-complex IR.

In contrast, it is possible to decouple the requirements of faithfully modeling Python from the requirements needed for transforms such as quantization or fusion. Transforms are often formulated as modifications to a high-level directed acyclic graph (DAG) organization of the code, with implementation details hidden within high-level blocks (such as Convolution or Batch Normalization). Thus, simplifications can be made to both the program capture mechanism and the IR it produces, focusing on the high-level DAG structure of the majority of neural network computation.

For this use case, we present \ic{torch.fx}, a high-productivity library for capturing and transforming PyTorch programs. \ic{torch.fx} explicitly trades generality of supported programs for simplicity of program capture and representation. \ic{torch.fx} focuses on the DAG representation of deep learning programs and provides customization interfaces to adapt programs into this representation. In doing so, \ic{torch.fx} is able to provide a program transform interface that supports the majority of deep learning programs while providing simple and easy-to-use APIs for implementing transforms.

We present the following contributions:
\begin{enumerate}
    \item A practical analysis of the features of program capture and transformation that are important for deep learning programs. 
    \item A Python-only program capture library that implements these features and can be customized to capture different levels of program detail.
    \item A simple 6 instruction IR for representing captured programs that focuses on ease of understanding and ease of doing static analysis.
    \item A code generation system for returning transformed code back to the host language's ecosystem.
    \item Case studies in how \ic{torch.fx} has been used in practice to develop features for performance optimization, program analysis, device lowering, and more.
\end{enumerate}

\section{Background}
\label{background}

\label{eagerdifferentiable}


When capturing and transforming programs, both eager and graph-mode frameworks must make choices about \emph{capturing program structure}, \emph{program specialization} and the \emph{design of the intermediate representation} in which programs are kept. The combination of these choices determines the space of programs that are representable in the framework, the ease of writing transformations, and the performance of resulting transformed programs. In general, supporting more programs at high performance requires a more complicated capture framework and IR and subsequently makes transformations harder to write.

\subsection{Capturing Program Structure}

There are several ways to capture program structure from Python programs. The simplest way is to \emph{trace} the execution of a model given some example inputs and record the operations that occur, which is the approach used by PyTorch's \ic{jit.trace}~\cite{torchscript}. A slightly more complicated variant of this approach is to perform tracing with abstract values rather than example inputs (\emph{symbolic tracing}). MXNet's Gluon~\cite{chen2015mxnet}, and TensorFlow's \ic{tf.function}~\cite{DBLP:journals/corr/abs-1810-08061} implement this approach. In addition to the user not having to provide example inputs, this approach surfaces locations where Python control flow depends on the input values, rather than collecting a trace specialized to the control decisions imparted by the example inputs.

During tracing, operations are only recorded for tensors and a small number of other data structures such as lists of tensors. This means that tracing can only record a representation for a subset of the Python program. Although tracing's visibility into the program is limited, this is often sufficient for deep learning computations, which are most often flat sequences of tensor operations\textemdash termed \textit{basic block} programs in Section \ref{ir_design}.

By overriding the execution behavior of standard Python code, some tracing systems can capture more program structure, such as control flow, at the cost of additional complexity. For instance, \ic{tf.function} augments symbolic tracing with a Lightweight Modular Staging~\cite{rompf2010lightweight} system that uses Python AST transforms to convert imperative control flow constructs into higher-order Python functions, which can then be traced.


An alternative way to capture program structure is to have users write models directly in an embedded programming language within Python. The simplest of these techniques is to provide a graph-building API similar to TensorFlow, which lets users build programs (graphs) by calling Python functions. It is awkward to represent control flow in these APIs, so PyTorch's TorchScript~\cite{torchscript} instead extracts programs directly from the Python source using a traditional lexer-parser-compiler toolchain. TorchScript can inspect the source syntax in full fidelity and can understand language constructs such as structured control flow, collection types (e.g. {\tt tuple}, {\tt list}, {\tt dict}) and user-defined types. As opposed to tracing, which can fail silently, embedded language approaches can report unsupported constructs as part of compilation. On the other hand, embedded language compilation is significantly more complicated to implement, since it requires a full language stack. Even then, in practice these systems will not support the full Python language, so users still need to make their program conform to the supported subset (albeit a larger subset than supported by tracing systems).

Systems such as Zygote.jl~\cite{DBLP:journals/corr/abs-1810-07951} and TPU integration~\cite{DBLP:journals/corr/abs-1810-09868} in the Julia ecosystem~\cite{bezanson2017julia} as well as Swift for TensorFlow~\cite{DBLP:journals/corr/abs-2102-13243} provide program transformation interfaces by way of integration into non-Python host languages. The main drawback of such native host language integrations in Swift and Julia is that they require the user to exit the Python ecosystem. Python has considerable momentum and extensive libraries in the numeric/scientific computing (and particularly deep learning) space, and many users prefer to stay in the Python ecosystem. While other languages may provide objectively better experiences in some respects, adoption has been slow.


\subsection{Specializing Programs}
\label{program_specialization}

A Python expression such as \ic{a + b} is very abstract. There are no constraints on the types of \ic{a} or \ic{b}. Even if both are Tensors, the number of dimensions and the size of the dimensions might vary. When ML frameworks capture programs, they often simultaneously \emph{specialize} these expressions such that they are only valid for specific types or tensor shapes. The more a program is specialized, the fewer inputs it will work on, so approaches vary in the \emph{degree} of specialization, the \emph{timing} of when specialization is done (ahead of time, just-in-time), and the \emph{safety} of the specialized result.

For example, PyTorch's TorchScript \ic{torch.jit.trace}~\cite{torchscript} specializes to the shape of the example inputs. \ic{jit.trace} capture is unintrusive\textemdash that is\textemdash it records the operations that occur during an actual execution run of the program. One implication of this is the presence of tensor metadata such as the \ic{ndim} or \ic{shape} attributes, which can escape the traced region and be used in control decisions within the Python program. This may cause the traced representation to be \emph{shape specialized}\textemdash that is\textemdash it is only valid for the value shapes used at trace time and may fail for other shapes.

To avoid the problem of specialization failing for some inputs, systems such as DyNet~\cite{neubig2017dynet} and LazyTensor~\cite{suhan2021lazytensor} perform tracing just-in-time, and thus can capture specialized program representations for every invocation. At runtime, these systems defer execution of tensor operations, instead accumulating a program trace. When a value must be materialized, the system will apply transformations to the collected program representation (e.g. automatic batching or native code lowering) and execute the code, returning the values requested. However, this process adds additional cost, since the program is captured on every invocation. LazyTensor uses a caching system to reduce this cost: optimized artifacts are stored in a cache keyed by a hash of the collected IR. On further invocations of the same IR, the optimized artifact can be called directly.

The performance of JIT specialization can also be improved by proving that re-capturing the program is unneeded for some inputs. For instance, JAX's \ic{jit} combinator~\cite{frostig2018compiling} uses pure, functional Python programs as input. This enforces referential transparency on non-Tensor computation like shape expressions. When some transform requires specialization, such as conversion to XLA~\cite{XLA} with static shapes, the system can look at the shapes of the inputs to determine if a new capture is required. A disadvantage of JIT specialization is that it is more complicated to reason about code execution. For instance, {\tt print} or {\tt pdb} statements in traced code will only be executed \emph{on runs where re-tracing occurs}. Re-tracing and re-transformation can also cause hard-to-predict performance bubbles as execution of the system stalls to re-specialize.


\subsection{Intermediate Representation Design}
\label{ir_design}

ML frameworks vary in the format of their IRs, with richer IRs capturing more programs and being more expressive at the cost of additional complexity to write transformations or run the code efficiently.

\paragraph{Language}  Many frameworks define their IR in a cross-language way. For example, Caffe and TensorFlow use the Protocol Buffers format~\cite{protobuf} to represent computational graphs. PyTorch's JIT and MXNet use C++ data structures for their IR with additional bindings into Python. Such native representations can have better runtime performance and may be easier to serialize. On the other hand, these representations can impose a learning curve above that required for programming Python.


\paragraph{Control flow} 
Most neural networks are expressible as flat sequences of tensor operations without control flow such as if-statements or loops\textemdash a definition we refer to as a \emph{basic block} program. \emph{Basic block} programs are often represented as a directed acyclic graph (DAG) data structure. Multilayer Perceptrons (MLPs), Convolutional Neural Networks (CNNs) such as ResNet~\cite{DBLP:journals/corr/HeZRS15} and personalization/recommendation models \cite{DBLP:journals/corr/abs-1906-00091} are easily expressed this way. Similarly, Transformer networks~\cite{DBLP:journals/corr/VaswaniSPUJGKP17} can also be expressed in this way, barring the loop needed for sequence generation on the decoder portion of the network.

Recurrent Neural Networks (RNNs) such as the Elman RNN~\cite{elman1990finding}, LSTM~\cite{hochreiter1997long}, and Gated Recurrent Unit (GRU)~\cite{DBLP:journals/corr/ChoMGBSB14} are not immediately expressible in this way, as the recurrent network computation is applied repeatedly across elements of a sequence with (typically) dynamic length. RNN structures can be represented in an imperative language as a loop with tensor computation applied in the loop body and tensor values carried across loop iterations. However, in practice, these RNN structures are typically provided as wholesale tensor operations. Thus, an entire RNN application over a sequence appears in code as a call to an RNN function or module. Therefore, these network architectures often also appear as \textit{basic block} programs.

Nevertheless, many frameworks support capturing and representing control flow in their IR. TorchScript built control flow support into all of its components first-class due to anticipation for workloads to become more complex, particularly in sequence processing domains. JAX uses higher-order functions such as \ic{jax.lax.scan} to allow functional-style control flow~\cite{frostig2018compiling}. MLIR represents control flow with basic blocks that end in tail calls~\cite{DBLP:journals/corr/abs-2002-11054}. In addition to adding complexity to the IR, more general control flow also makes transforms such as common sub-expressions more complicated to implement.



\paragraph{State} Deep learning models contain state in the form of the trainable model weights used in different layers. Apart from these parameters, most networks operate as pure functions of their inputs. ML frameworks take different approaches to handling how this state is mutated.

PyTorch allows values to be mutated and tensors can be views of each other. For example, the slicing syntax \ic{x[i]} (where \ic{x} is a Tensor value) does not produce a new Tensor value, but rather returns a view aliasing the subset of tensor \ic{x} indexed by \ic{i}. Views can also be mutated. For example, the expression \ic{x[i] = y} will write the value of \ic{y} into the portion of \ic{x} indexed by \ic{i}. 

Since PyTorch supports these aliasing and mutation semantics, modifications to programs must be done in the context of an analysis that proves that the modification is safe~\cite{andersen1994program}. TorchScript implemented such alias analysis for the purpose of reasoning about the safety of transforms over the TorchScript IR. However, this comes at a high cost: all operations in the program must be annotated with information specifying their aliasing and mutation behavior. In practice, many functions (opaque calls or ones that have not been annotated with relaxed semantics) are treated with a conservative assumption that the callee mutates global memory, causing the operation to act as a barrier and hindering optimization. Needing to reason about aliasing and mutability complicates pass authoring, adds additional maintenance burden to the framework, and can limit optimization opportunities, but enables the user to apply the full generality of the PyTorch tensor language.

JAX's functional approach moves the burden of tracking this state outside of the framework. Instead the model must be turned into a pure function where the parameters are passed as inputs. Typically, this is done with wrapper libraries such as Haiku~\cite{haiku2020github} or Flax~\cite{flax2020github}. Any transforms that have to modify both state and code, such as folding batch norm scaling to a weight tensor, are made more complicated because these components no longer live together in the same framework.


\section{Design Principles}
Many of the different designs for program capture and transformation used in existing frameworks favor the ability to represent more deep learning programs at the cost of the complexity of their implementation. When captured programs are the \emph{only} way to run a program, the ability to capture a program in full fidelity is crucial. But PyTorch is primarily used as an \emph{eager execution} framework and program capture is only used for some specific transforms; It does not need to work for an entire program. Furthermore, most PyTorch programmers who want to transform models are machine learning practitioners who prefer to work in Python and may have less knowledge of compiler design.

By designing for \emph{typical} deep learning models rather than the long tail, it is possible to create a framework that is much easier to use and simpler to implement. This philosophy is captured by \ic{torch.fx}'s design principles:

\begin{itemize}
    \item Prefer making program capture and transformation easy for typical models at the cost of working for all possible programs. Avoid complexity to support long-tail, esoteric use cases.
    \item Work with tools and concepts that ML practitioners are already familiar with such as Python data structures and the publicly documented operators in PyTorch.
    \item Make the process of program capture highly configurable so users can implement their own solutions for long-tail uses. Allowing users to make one-off configurations is simpler than handling the general case.
\end{itemize}

\section{torch.fx Overview}
\label{fx_overview}

In the spirit of simplicity, \ic{torch.fx} \emph{captures programs} via symbolic tracing, \emph{represents them} using a simple 6-instruction python-based IR, and \emph{re-generates Python code} from the IR to execute it. To avoid the complexities of re-capture for JIT specialization, \ic{torch.fx} makes no attempt to specialize programs itself, instead relying on the transforms to decide what specializations they want to perform during capture. The process of symbolic tracing can be configured by users to work for more esoteric uses.

\begin{figure}[!t]
\begin{lstlisting}
import torch
from torch.fx import symbolic_trace, GraphModule

def my_func(x):
  return torch.relu(x).neg()

# Program capture via symbolic tracing
traced : GraphModule = symbolic_trace(my_func)
for n in traced.graph.nodes:
  print(f'{n.name} = {n.op} target={n.target} args={n.args}')
"""
x = placeholder target=x args=()
relu = call_function target=<built-in method relu ...> args=(x,)
neg = call_method target=neg args=(relu,)
output = output target=output args=(neg,)
"""

print(traced.code)
"""
def forward(self, x):
    relu = torch.relu(x);  x = None
    neg = relu.neg();  relu = None
    return neg
"""
\end{lstlisting}
\caption{\ic{torch.fx} captures programs using symbolic tracing into a simple IR and generates Python code from that IR.}
\label{fx_components_code}
\end{figure}

Figure \ref{fx_components_code} shows an example of capturing code with \ic{torch.fx}. \ic{symbolic_trace} takes a function or \ic{torch.nn.Module} and captures its structure in a \ic{Graph} object. That \ic{Graph} object is combined with module parameters in a \ic{GraphModule}, which is a subclass of \ic{torch.nn.Module} whose \ic{forward} method runs the captured \ic{Graph}. We can print the \ic{Node}s of this \ic{Graph} to see the IR that was captured. \ic{placeholder} nodes represent inputs and a single \ic{output} node represents the result of the \ic{Graph}. \ic{call_function} nodes have a reference directly to the Python function they would call. \ic{call_method} nodes directly invoke a method on their first argument. The \ic{Graph} is reconstituted into Python code (\ic{traced.code}) for invocation.

\begin{figure}[!t]
    \begin{lstlisting}
from torch.fx import Graph
def replace_activation(g: Graph, old, new):
  for n in g.nodes:
    if n.op == 'call_function' and n.target == old:
       # create IR to call new activate
       with g.inserting_after(n):
         new_n = g.call_function(new, n.args)
         n.replace_all_uses_with(new_n)
         g.erase_node(n)
       # or for this simplified case: `n.target = new` 
    
replace_activation(traced.graph, torch.relu, 
                   torch.nn.functional.gelu)
traced.recompile()
    \end{lstlisting}
    \caption{\label{transform_example} Transforms, like this one that replaces activation functions, are written directly in Python.}
\end{figure}

Figure \ref{transform_example} shows an example transform using \ic{torch.fx}. The transform finds all instances of one activation and replaces them with another. We use it replace \ic{relu} with \ic{gelu} in our example.

\subsection{Program Capture}
\label{program_capture_symtrace}

\ic{torch.fx}'s symbolic tracing mechanism uses a \ic{Proxy} data structure to record operations on values flowing through the program. \ic{Proxy} is a duck-typed Python class that records attribute accesses and method calls on it, acting as an abstract value that stands in for the concrete program values. \ic{Proxy} uses the \ic{__torch_function__} protocol~\cite{torch_function} to intercept and record the dispatch of PyTorch operators, which are free functions. Finally, \ic{torch.fx} overrides PyTorch's \ic{Module} abstraction to record calls to \ic{Module}s using proxied values. The process of symbolic tracing is configurable via a \ic{Tracer} class whose methods can be overridden to control what values are kept as \ic{Proxy}s and which are partially evaluated during the trace. 

\subsection{Intermediate Representation}
\label{ir_what}

\ic{torch.fx} represents programs in a DAG-based IR, which is amenable to the \emph{basic block} programs common in deep learning. Programs are represented as a \ic{Graph} object, which contains a linear series of \ic{Node} objects representing operations. \ic{Node}s have a string opcode, describing what type of operation the \ic{Node} represents (the semantics of the opcodes can be found in Appendix \ref{node_semantics}). Nodes have an associated \ic{target}, which is the call target for call nodes (\ic{call_module}, \ic{call_function}, and \ic{call_method}). Finally, \ic{Node}s have \ic{args} and \ic{kwargs}, which together represent the arguments to the target in the Python calling convention as witnessed during tracing\footnote{No normalization is applied to args or kwargs; They are preserved as the user wrote them. This facilitates further backward-compatibility of the generated code} (the semantics for \ic{args} and \ic{kwargs} for each opcode can be found in Appendix \ref{args_kwargs_semantics}). Data dependencies between \ic{Node}s are represented as references to other \ic{Node}s within \ic{args} and \ic{kwargs}.

To simplify the IR, \ic{torch.fx}'s IR does not have primitive operations that model the construction or mutation of data structures. Nevertheless, \ic{args} and \ic{kwargs} support immediate values: Python built-in types such as \ic{int} and \ic{float} and recursive collection types like \ic{tuple} and \ic{list} can appear as \ic{Node} arguments without separate object construction \ic{Node}s. Because \ic{Node}s support immediate values, the IR is clean and \ic{Node}s are approximately 1-to-1 with Tensor operations.

\ic{torch.fx} stores the state of the program in the \ic{GraphModule} class. \ic{GraphModule} is the container for transformed programs, exposing the transformed, generated code as well as providing the familiar parameter management APIs of \ic{nn.Module}. \ic{GraphModule} can be used anywhere a normal \ic{nn.Module} can be used, providing interoperability between transformed code and the rest of the PyTorch ecosystem. 

\ic{torch.fx}'s IR provides two opcodes for accessing state in the \ic{Module} hierarchy: \ic{call_module}, which invokes a sub-\ic{Module}'s forward method, and \ic{get_attr}, which fetches a parameter from the \ic{Module}. Transformed code can interact with the \ic{Module} hierarchy in much the same way normal PyTorch code can via these opcodes. In addition, transformations can manipulate the mutable state in the \ic{Module} hierarchy simultaneously with transformations over code. This provides a natural separation between the mutable parameters and the functional \ic{Graph} that interacts with them via \ic{call_module} \ic{Node}s, while still keeping them together in a single object for doing transformations that work on both.

\subsection{Source-to-Source Transformation}
\label{codegen_what}

The final stage in the \ic{torch.fx} transformation pipeline is code generation. Rather than exiting the Python ecosystem and entering a bespoke runtime, \ic{torch.fx} generates valid Python source code from the transformed IR. This transformed code is then loaded into Python, producing a callable Python object, and installed as a \ic{forward} method on the \ic{GraphModule} instance. Using code generation allows the results of \ic{torch.fx} transforms to be installed in models and still used in further transforms. For instance, in Figure \ref{code_generation_example} we take the result of tracing our original program and install it as the activation in a new module. Then, we symbolically trace the result for further transformation. 

\begin{figure}[!t]
    \begin{lstlisting}
class SampleModule(torch.nn.Module):
    def forward(self, x):
      return self.act(x + math.pi)
  
sm = SampleModule()
sm.act = traced # from previous figure
traced : GraphModule = symbolic_trace(sm)
print(traced.code)
"""
def forward(self, x):
    add = x + 3.141592653589793;  x = None
    gelu = torch.nn.functional.gelu(add);  add = None
    neg = gelu.neg();  gelu = None
    return neg
"""
    \end{lstlisting}
    \caption{\label{code_generation_example} \ic{torch.fx} generates Python code as its output, so it can be reused in further capture and transform steps.}
\end{figure}

\section{Design Decisions}
\label{designdecisions}

\ic{torch.fx} mixes and extends approaches from previous work to deliver an easy to use, simple to implement, and configurable library. We highlight a few of these decisions here.

\subsection{Symbolic Tracing}

\ic{torch.fx} uses symbolic tracing with \ic{Proxy} objects rather than embedded language techniques because they are easier to implement directly in Python using its flexible object model. The implementation is simple enough that users can read and step through the source when tracing behaves unexpectedly.

Tracing also helps eliminate control flow in a model not dependent on inputs such as the loop over sequential modules in a \ic{torch.nn.Sequential}. PyTorch models are written pervasively with these abstractions, with many users also using third party libraries that contain their own model implementations, so it is important to be able to trace through these abstractions to get to the actual operators running.

Symbolic tracing works well for common models at the cost of not being able to capture long-tail models that actually contain input-dependent control flow. We make up for this limitation by making the tracing process customizable to work around one-off issues.

\subsection{Configurable Program Capture}
\label{configurable_program_capture}

\ic{torch.fx}'s symbolic tracing is customizable. A \ic{Tracer} class controls the behavior of \ic{fx.symbolic_trace}. Its methods can be overridden to change the tracing process's behavior.

The \ic{is_leaf_module} method can be overridden to specify which PyTorch \ic{Module} instances should be treated as opaque calls during tracing. By default, \ic{torch.fx} keeps PyTorch built-in \ic{Module}s such as \ic{nn.Conv2d} intact while tracing through user-defined \ic{Module}s, since this creates a trace of standard, understandable primitives.  Customizing this behavior can block out portions of a model that contain unsupported language features or modify the level of representation used for transformations. 

\ic{create_proxy} is a method that can be overridden to customize the behavior of creating a \ic{Node} in the \ic{Graph} and the associated runtime \ic{Proxy} value. This can be used to, for example, install custom metadata onto \ic{Node}s for the purpose of transformation or to support custom data structures as traceable values. A custom \ic{Tracer} could, for instance, specialize the sizes and shapes of \ic{Tensor}s and use these values to capture a program that would otherwise not be traceable without specialization.

\subsection{AoT Capture without Specialization}
\label{aot}

While ahead-of-time tracing limits the space of programs that can be captured (e.g. arbitrary control flow is not supported), it provides a more predictable and more observable capture, transformation, and code generation process that fits into the PyTorch developer experience and works well in practice.

Unlike example-based tracing, symbolic tracing cannot incidentally specialize program flow because the information needed to make data-dependent control flow decisions is not present at trace time. Common \ic{Tensor} attributes used in control decisions such as \ic{shape} and \ic{ndim} are returned as \ic{Proxy} values during symbolic tracing. Operations on these values can then be recorded. On the other hand, when these \ic{Proxy} objects are used in a context where untraceable operations (such as a cast to Python built-in types like {\tt int} or {\tt bool}) occur on them, the user receives an error message describing the problem and a stack trace indicating the location of the issue.

\subsection{Python-based IR and Transforms}
\label{codegen}

Rather than use a cross-language format such as protocol buffers, \ic{torch.fx} IR is entirely represented and implemented Python. Users can call, read, or override it easily. There is no need to understand Protocol Buffers or C++ (or set up either of their build environments), which present barriers to ML engineers familiar with working primarily in Python. Transforms are written in Python as well. 

Furthermore, the \emph{result} of transformations is also Python code. This makes it easy to inspect for correctness, debug with \ic{pdb}, feed to libraries, and pass on to further transforms. Transformed code is encapsulated in a \ic{GraphModule} that can be used in PyTorch just like any other \ic{nn.Module}. For instance, a user can TorchScript compile the model for deployment or use it in PyTorch's \ic{DistributedDataParallel} library. Users can also save the generated code as a source file via the experimental \ic{GraphModule.to_folder} API.

Code generation further integrates \ic{torch.fx} into the Python ecosystem rather than sequestering transformed code into a bespoke and harder-to-use runtime.

\subsection{No Control Flow Inside IR}
\label{controlflow}

With Transformers~\cite{DBLP:journals/corr/VaswaniSPUJGKP17} increasingly replacing sequential recursive neural networks with larger scalable attention modules, the use of host language control flow in deep learning is becoming more rare. Many models can be expressed without it, and even for programs with some control flow (e.g. a beam search decoder), there are large blocks of the model without control flow (the encoder and the step of the decoder).

However, the presence of control flow in an IR adds significant complexity regardless of whether a particular model uses it. Most analyses on the IR must be expressed as fix-point data-flow~\cite{kildall1972global} over the program rather than simple forward propagation.  The author must define a lattice, transfer function, and join function for the analyzed property in the program and prove monotonicity and finiteness thereof. While familiar to compiler writers, we have found that writers of ML transforms often introduce bugs in transforms such as having join functions that are not monotonic or failing to iterate until converged. In contrast, for a \emph{basic block} IR, only a transfer function is needed.

An example of the complexity of fix-point analysis can be found in shape propagation: shapes can be trivially propagated forward through a basic block program (barring a few operations with value-dependent output shapes). However, when control flow is added, shape propagation does not satisfy the finiteness property\textemdash a value carried across a loop iteration can take on an infinite number of shapes, as shown in Figure~\ref{lcd_shape_analysis}. The analysis will typically reach a ``dynamic" value in such situations. Shape analysis would then provide under-specified data, which would hinder further transformations that require concrete shape information, such as ASIC lowering.

\begin{figure}[!ht]
    \begin{lstlisting}
def loop_shapes(x, itr):
  # x is an input tensor of size [1, N]

  for _ in range(itr):
    x = torch.cat((x, x), dim=0)

  # Depending on the number of loop iterations, x may have an
  # arbitrary leading dimension i.e. x \in [*dynamic*, N]
  return x
    \end{lstlisting}
    \caption{\label{lcd_shape_analysis} A demonstration of dynamic shapes due to loop-carried dependencies}
\end{figure}

Furthermore, some transformations proposed in the ML community are not well defined in the presence of control flow, such as the quantization transform described in Section \ref{quant}.

The fact that the IR does not contain control flow itself does not prevent transforms from working on sub-graphs of basic blocks within a larger model; We leave the details of how this composition works to the writer of the transform or the user applying the transform.

\subsection{Functional Graphs but Stateful Modules}

As described in Section~\ref{ir_design}, aliasing and mutability semantics in a language can necessitate complex analyses to prove that a program transformation is legal. \ic{torch.fx} omits such analysis, instead defining mutating operations as undefined behavior with the option to raise errors when it is captured during tracing.

Avoiding mutability in the IR simplifies analysis and transformation of deep learning programs greatly. Most models do not suffer from this restriction since most mutation is localized to the parameters of the model.

\ic{torch.fx} still preserves the hierarchical \ic{nn.Module} structure from PyTorch and can represent module calls and attribute fetches from this structure. Modules like \ic{torch.nn.Conv2d} are well understood by users, have well-documented arguments, and hide the stateful use of parameters within the module, so preserving these objects makes writing transformations easier. For instance, a \ic{torch.nn.BatchNorm} module will actually contain mutable state, but that state is well understood by ML practitioners.

\section{Case Studies and Evaluation}
\label{case_studies}

\ic{torch.fx} has been used by PyTorch users both in the open-source ecosystem as well as as a critical component of the deep learning stack at a major software company. We study the complexity of \ic{torch.fx}'s IR and various use cases of \ic{torch.fx}, including \emph{performance optimization}, \emph{program analysis}, and \emph{device and runtime export}.

\subsection{IR Complexity}

One of the goals of \ic{torch.fx} is to simplify the IR produced for ML models and make it easier for ML practitioners to understand. We can compare \ic{torch.fx} IR to the IR produced by the two TorchScript~\cite{torchscript} front-ends (\ic{jit.trace} and \ic{jit.script}), since all start from the same input programs. Figure~\ref{ts_ir_complex} shows some example IR from the start of a ResNet model. The IR produced by TorchScript is very rich, including tensor operations, scalar operations, control flow, data structures, hierarchical module structure, and aliasing and mutability semantics. Support for these features makes it much more verbose for simple models, resulting in 2614 operations from \ic{jit.script} and 860 from \ic{jit.trace}. The same ResNet model consists of 445 operations in \ic{torch.fx} IR. Most of the reduction comes from eliminating control flow irrelevant to the captured trace. But \ic{torch.fx} IR also benefits from inlining simple constants and data structures, so is almost half the size of the IR the captured with \ic{torch.jit.trace}, which similarly eliminates control flow.

The complex IR from the TorchScript front-ends induces complexity in the program transform authoring process, requiring more care to write transforms correctly and leading to longer and less maintainable transform code. \ic{torch.fx} addresses this by greatly simplifying its captured representation, facilitating transforms that are easier to write and maintain.







\begin{figure*}[t]
     \begin{subfigure}[b]{.5\textwidth}
     \begin{lstlisting}
graph(%self : torchvision.models.resnet.ResNet,
      %x.1 : Tensor):
  %13 : str = prim::Constant[value="AssertionError: "]()
  %14 : bool = prim::Constant[value=0]() 
  %15 : float = prim::Constant[value=1.0000000000000001e-05]() 
  %16 : float = prim::Constant[value=0.10000000000000001]()
  %17 : str = prim::Constant[value="..."]() 
  %19 : bool = prim::Constant[value=1]() 
  %20 : int = prim::Constant[value=2]() 
  %21 : int = prim::Constant[value=3]() 
  %23 : int = prim::Constant[value=-1]()
  %24 : ...Conv2d = prim::GetAttr[name="conv1"](%self)
  %25 : Tensor = prim::GetAttr[name="weight"](%24)
  %26 : Tensor? = prim::GetAttr[name="bias"](%24)
  %27 : int[] = prim::ListConstruct(%20, %20)
  %28 : int[] = prim::ListConstruct(%21, %21)
  %29 : int[] = prim::ListConstruct(%19, %19)
  %x.5 : Tensor = aten::conv2d(%x.1, %25, %26, %27, %28, %29, %22) 
  ...
\end{lstlisting}
         \caption{{\scriptsize TorchScript IR}}
         \label{example_export}
     \end{subfigure}
     \begin{subfigure}[b]{.5\textwidth}
\begin{lstlisting}
def forward(self, x : torch.Tensor) -> torch.Tensor:
    conv1_weight = self.conv1.weight
    conv2d = torch.conv2d(x, conv1_weight, None, 
                          (2, 2), (3, 3), (1, 1), 1)
    ...
\end{lstlisting}
         \caption{{\scriptsize \ic{torch.fx} IR}}
         \label{importing_api}
     \end{subfigure}
     
        \caption{\ic{torch.fx} traces through non-varying control flow and can embed constants as arguments in its \ic{Node}s. This substantially simplifies the IR for typical models.
        For a canonical ResNet50 model, \ic{torch.fx} IR contains 445 operations compared to 2614 for \ic{torch.jit.script} and 860 for \ic{torch.jit.trace}.}
        \label{ts_ir_complex}
\end{figure*}

\label{eval_fx_ir_complexity}

\subsection{Performance Optimization}
\label{perfoptimization}

PyTorch's tensor language provides good performance in many cases, but architectural details of the underlying hardware create opportunities for further optimization. We investigate techniques by which \ic{torch.fx} enables runtime performance improvements.

\subsubsection{Quantization}
\label{quant}

Quantization~\cite{DBLP:journals/corr/abs-1712-05877} is a technique used to increase the efficiency of neural network computation by reducing the size of Tensor data elements. Smaller data elements require less memory bandwidth, less storage, and can often be processed faster by modern processors. Neural network computation has relaxed sensitivity to numerical perturbations, so quantization is a canonical performance optimization.

Performing Post-Training Quantization or Quantization-Aware Training requires access not only to parameter values but also to the activation values that flow through the program~\cite{DBLP:journals/corr/abs-1806-08342}. For instance, quantization-aware training needs to measure the distribution of floating point values in the output of a tensor addition operation to calculate a scale and bias value under quantized numerics. Such introspection is generally not available in PyTorch eager mode. However, \ic{torch.fx} provides a lightweight way to capture such a program representation.

The Post-Training Quantization procedure entails the following stages:

\begin{enumerate}
    \item A preparation phase, which instruments the program with ``observer" objects that record statistical information about the floating-point values contained in Tensor values at various points in the program.
    \item A calibration phase, where the user feeds batches of data through the network to populate the observers.
    \item A conversion phase, where the collected statistics are used to down-cast weight values and convert operations in the model to quantized operations with embedded scale and zero-point information.
\end{enumerate}

Quantization makes use of \ic{torch.fx}'s graph and \ic{GraphModule} representation to simultaneously modify the program code and weight values. The process for Quantization-Aware Training is analogous to phases (1) and (2) in the above but with ``fake quantize" observers that snap floating point values to the corresponding values under quantized numerics. 

\begin{figure}[!ht]
    \resizebox{\linewidth}{!}{
        \includegraphics[]{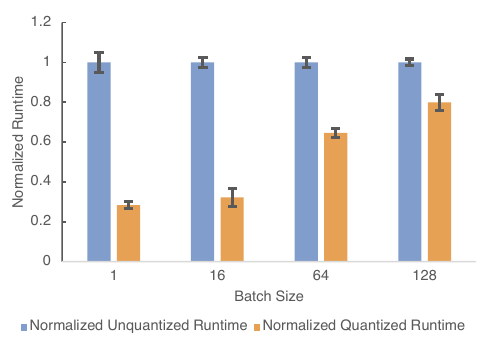}
    }
    \caption{\label{quant_exp_results_inline} Normalized inference runtime (lower is better) for \ic{torch.fx}-based quantization.}
\end{figure}

We evaluate the performance of a DeepRecommender~\cite{kuchaiev2017training} model with Post-Training Quantization applied on a server-class Intel Xeon Gold 6138 CPU @ 2.00GHz using FBGEMM~\cite{DBLP:journals/corr/abs-2101-05615} quantized operations. Figure~\ref{quant_exp_results_inline} shows that \ic{torch.fx}-enabled quantization confers up to a 3.3x runtime performance improvement compared to the floating point model, with low variance highlighting the predictable performance characteristics of ahead-of-time transformation. Correctness testing of quantization is not straightforward since it is a semantics-changing transform, but the applicability of numerics on this workflow has been validated on several model architectures via evaluation set testing. Numeric data for the experiment can be found in Appendix \ref{quant_data}. The preparation phase takes 44 ms, the calibration phase takes 590 ms, and the conversion phase takes 3.8 seconds. The majority of the time in the latter two phases can be attributed to tensor operations during model execution or value quantization, respectively.

Not only does \ic{torch.fx}-based quantization provide the expected performance increases, but the tool's development saw an order-of-magnitude productivity increase compared to an implementation on the TorchScript platform. By reducing the amount of complexity in the representation, exposing transformation APIs in Python, and embedding into the native PyTorch ecosystem, \ic{torch.fx} provides a high-productivity environment for semantics-changing transforms like quantization.

\subsubsection{Fusion Optimizations}

Operator Fusion is a class of optimization that merges patterns of tensor operations together into a single compute kernel. Fusion can save operator dispatch cost, memory bandwidth cost, and memory space cost.

One example of operator fusion is \textit{Convolution-BatchNorm fusion}. During inference, a Convolution-BatchNorm operator sequence can be merged by applying the batch normalization weights to the convolution weights~\cite{markus_2018}.

\begin{figure}[!ht]
    \includegraphics[]{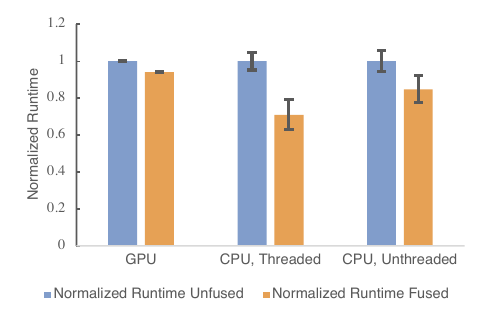}
    \caption{\label{conv_bn_fusion_results} Normalized inference runtime (lower is better) with \ic{torch.fx}-based Convolution/Batch-Norm fusion.}
\end{figure}

We evaluate this transformation on a PyTorch ResNet50 model on an NVIDIA Tesla V100-SXM2 16GB with CUDA version 11.0 and an Intel Xeon Gold 6138 CPU @ 2.00GHz. Figure \ref{conv_bn_fusion_results} shows approximately a 6\% latency reduction for the GPU case, a 40\% latency reduction on CPU with default intra-op parallelism, and a smaller 18\% latency reduction with intra-op parallelism disabled (i.e. \ic{OMP_NUM_THREADS=1}). Numerical correctness is confirmed via an epsilon equivalence comparison (\ic{rtol=1e-05}, \ic{atol=1e-08}) of the outputs of the fused and unfused implementations. Numeric results for this experiment can be found in Appendix \ref{fusion_data}. The runtime of the transformation itself was 81 ms, the majority of which consists of the arithmetic operations to fuse the parameter tensors together.

\ic{torch.fx} provides the necessary non-local program context and state modification facilities needed for this transformation with its ahead-of-time, graph-based nature~\cite{he_2021}. The whole transformation and test harness amount to fewer than 150 lines of Python, demonstrating the power of \ic{torch.fx}'s APIs in enabling concise, fast-to-develop program transformations over PyTorch code.

\subsubsection{Program Scheduling}


Large PyTorch models sometimes contain blocking remote procedure calls to fetch values from parameter servers. For clarity these calls are written right before the parameters are used. However if a model contains several such calls, better utilization is achieved by overlapping these networks calls with other local work. With \ic{torch.fx}, we provide a pass that replaces the blocking network calls with non-blocking ones and a separate wait call. We then hoist the non-blocking call as early as possible in the program. On large distributed training jobs, we have found this optimization can increase QPS by up to 9\%.

\subsection{Program Analysis}

\ic{torch.fx} has been applied in various ways for program analysis.

\ic{torch.fx} has been used to implement a framework for simulation of deep learning inference at scale on various hardware devices at a major software company. \ic{torch.fx} enables the estimation of FLOPs, memory bandwidth usage, and data value sizes of the workload, allowing for estimation of the program runtime and memory consumption. This system allows for rapid development of deep learning systems, enabling quick iteration in simulation rather than on real devices.

\ic{torch.fx} has also been used for various forms of shape analysis. The canonical \ic{fx.passes.shape_prop} package provides a naïve implementation of shape analysis by interpreting the graph and recording the observed shapes. Additional systems, including shape propagation via symbolic expressions and shape propagation via gradual typing semantics, are in development. \ic{torch.fx} provides a representation on which such analyses can be done, opening opportunities for type system and inference innovations to be applied to PyTorch models.

Finally, \ic{torch.fx} provides an \ic{fx.graph_drawer} package, which gives the user the ability to visualize \ic{torch.fx} graphs with Graphviz~\cite{10.1007/3-540-45848-4_57}. This provides a commonly-requested way of understanding a deep learning program via a visual representation of its DAG.

\subsection{Device and Runtime Export/Compilation}

PyTorch is primarily designed for modern GPUs, which provide a great deal of flexibility and dynamism and thus are very amenable to PyTorch's \emph{eager mode} execution model. However, GPUs can still benefit from ahead-of-time compilation of model code through tookits like NVIDIA's TensorRT~\cite{tensorrt}.

More specialized processors (such as the TPU~\cite{jouppi2017datacenter}) promise higher performance, better power efficiency, and reduced cost via specialized functional units, specialized number formats, and new memory architectures. These processors often require static analyses and optimizations including operator scheduling, code generation, memory planning/scheduling, and architecture-aware quantization. Similarly to the optimizations in \ref{perfoptimization}, such analyses typically require greater program context than the per-operator kernel launches provided by PyTorch during eager mode execution. \ic{torch.fx} provides a pathway for such compiler stacks to integrate with PyTorch by providing a program representation extracted ahead-of-time. \ic{torch.fx} is used at a major software company for ASIC lowering.

\begin{figure}[!ht]
    \includegraphics[]{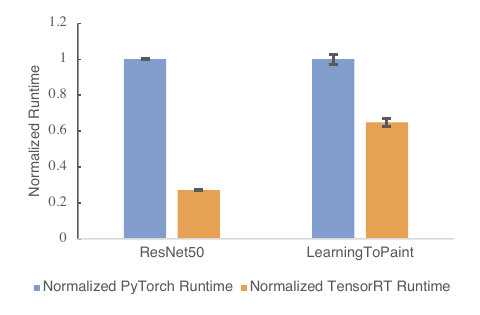}
    \caption{\label{trt_results} Normalized inference runtime (lower is better) with \ic{torch.fx}-based TensorRT lowering}
\end{figure}

We evaluate lowering a PyTorch ResNet50 model and a LearningToPaint model \cite{DBLP:journals/corr/abs-1903-04411} to NVIDIA TensorRT on an NVIDIA Tesla V100-SXM2 16GB GPU with CUDA version 11.0 using an experimental \ic{torch.fx}-to-TensorRT lowering system. Figure \ref{trt_results} shows that TensorRT provides a predictable 3.7x runtime speed-up across 30 trials compared to baseline PyTorch for ResNet50 and a 1.54x speed-up for LearningToPaint. Numerical correctness is confirmed via an epsilon equivalence comparison (\ic{rtol=1e-05}, \ic{atol=1e-08}) of the outputs of the TensorRT and non-TensorRT implementations. Numerical data for this experiment is available in Appendix \ref{trt_data}.

In addition to providing the platform for runtime speed-up through TensorRT, \ic{torch.fx} also provided high developer productivity for this component. The project was quickly developed using \ic{torch.fx}'s Python APIs as well as TensorRT's Python APIs, creating a translation layer between the two. The project was also able to quickly build components such as automatic splitting of the model based on TensorRT's supported operators and automatically scheduling unsupported operations in non-optimized blocks. Finally, the ultimate user API is very easy to use, inspect, and debug, as it conforms to Python coding practices.

\section{Conclusion}
\label{conclusion}

We presented \ic{torch.fx}, a Python-only system for capturing and transforming PyTorch programs. We analyzed the factors that complicated related systems\textemdash including control flow, mutability, and data model\textemdash and show how \ic{torch.fx} avoids complexity by focusing on common use cases and customizability. We investigated various use cases of \ic{torch.fx} across optimization, analysis, and device lowering, and show how these results are enabled by \ic{torch.fx}'s API design.


\section{Acknowledgements}

We would like to acknowledge all of the contributors to the core \ic{torch.fx} framework, including Alban Desmaison, Alex Beloi, Alexander Soare, Allen (Congcong) Chen, Andrew Millspaugh, Ansley Ussery, Aravind Kalaiah, Bradley Davis, Brandon Lin, David Esiobu, Dmytro Dzhulgakov, Eli Uriegas, Erjia Guan, Garret Catron, Harut Movsisyan, Horace He, Hui Guo, James Reed, Jason Ansel, Jay Leverett, Jerry Cai, Jerry Zhang, Jordan Fix, Kefei Lu, Lu Fang, Malay Bag, Meghan Lele, Mehdi Mirzazadeh, Michael Benayoun, Michael Suo, Mike Ruberry, Mikhail Zolotukhin, Natalia Gimelshein, Nikita Shulga, Oleg Khabinov, Onyiee, Patrick Hu, Patrick Spencer, Peter Bell, Philip Meier, Richard Zou, Sam Estep, Shirong Wu, Shiyan Deng, Thomas Wang, Vasiliy Kuznetsov, Yinghai Lu, Zachary DeVito, and Zeina Migeed.

We would like to acknowledge the contributors to the FX Graph Mode Quantization project used in evaluations, including Adnios, Alban Desmaison, Angela Yi, Bradley Davis, Charles David Hernandez, Emily Shen, Erjia Guan, Horace He, James Reed, Jerry Zhang, Raghuraman Krishnamoorthi, Mike Ruberry, Mikhail Zolotukhin, Philip Meier, Rong Rong (AI Infra), Sam Estep, Supriya Rao, Vasiliy Kuznetsov, Xiang Gao, Zachary DeVito, and Zafar Takhirov.

We would like to acknowledge the contributors to the fx2trt project used in evaluations, including Alex Beloi, Aravind Kalaiah, Bangsheng Tang, Eli Uriegas, Emad El-Haraty, Ivan Kobzarev, Jack Montgomery, James Reed, Jerry Zhang, Jordan Fix, Kefei Lu, Linbin Yu, Marat Subkhankulov, Mike Ruberry, Mor Tzur, Nikita Shulga, Philip Meier, Protonu Basu, Rui Zhu, Samuel Salas, Shirong Wu, Shiyan Deng, Yinghai Lu, and Zhengxu Chen.

Finally, we'd like to acknowledge all the discussions and feedback from many users inside and outside Facebook.





\bibliography{main}
\bibliographystyle{mlsys2022}

\appendix

\section{torch.fx Node Semantics}

\subsection{Opcode Meanings}
\label{node_semantics}

\begin{tabularx}{\linewidth}{|l|X|}
    \hline
     \textbf{Opcode} & \textbf{Meaning} \\
    \hline
     \ic{placeholder} & Function Input \\
    \hline
     \ic{call_method} & Call method on \ic{args[0]} \\
    \hline
     \ic{call_module} & Call module specified by \ic{target} \\
    \hline
     \ic{call_function} & Call function specified by \ic{target} \\
    \hline
     \ic{get_attr} & Retrieve attribute specified by \ic{target} \\
    \hline
     \ic{output} & Return statement; return \ic{args[0]} \\
    \hline
\end{tabularx}

\subsection{\ic{args}/\ic{kwargs} Behavior}
\label{args_kwargs_semantics}
\begin{tabularx}{\linewidth}{|l|X|}
    \hline
     \textbf{Opcode} & \textbf{\ic{args}/\ic{kwargs} Behavior} \\
    \hline
     \ic{placeholder} & Empty or \ic{args[0]} = default value \\
    \hline
     \ic{call_method} & Python calling convention; \ic{args[0]} is self \\
    \hline
     \ic{call_module} & Python calling convention; \ic{target} is self \\
    \hline
     \ic{call_function} & Python calling convention; \ic{target} is self \\
    \hline
     \ic{get_attr} & Empty \\
    \hline
     \ic{output} & \ic{args[0]} is the return value \\
    \hline
\end{tabularx}

\section{Quantization Evaluation Numeric Data}
\label{quant_data}

\resizebox{\linewidth}{!}{
\begin{tabularx}{1.1\linewidth}{|X|X|X|X|X|}
    \hline
     \textbf{Batch Size} &
     \textbf{Runtime Unquantized} &
     \textbf{stdev Unquantized} &
     \textbf{Runtime Quantized} &
     \textbf{stdev Quantized} \\
    \hline
     1 & 0.0777 & 0.00079 & 0.0222 & 0.0008 \\
    \hline
     16 & 0.1980 & 0.0104 & 0.0639 & 0.0057 \\
    \hline
     64 & 0.3995 & 0.0204 & 0.2585 & 0.0129 \\
    \hline
     128 & 0.6717 & 0.0228 & 0.5369 & 0.0413 \\
    \hline
     256 & 1.2307 & 0.0874 & 1.1157 & 0.0686 \\
    \hline
\end{tabularx}}

\section{Fusion Evaluation Numeric Data}
\label{fusion_data}

\begin{tabularx}{\linewidth}{|X|X|l|X|X|X|}
    \hline
     \textbf{Device} & \textbf{Fusion} & \textbf{Threads} & \textbf{Average runtime (sec)} & \textbf{stdev runtime} \\
    \hline
     GPU & Unfused & N/A & 0.1887 & 0.00048 \\
    \hline
     GPU & Fused & N/A & 0.1777 & 0.00049 \\
    \hline
     CPU & Unfused & Threaded & 0.2996 & 0.02835 \\
    \hline
     CPU & Fused & Threaded & 0.2129 & 0.03491 \\
    \hline
     CPU & Unfused & Unthreaded & 2.0231 & 0.23050 \\
    \hline
     CPU & Fused & Unthreaded & 1.7166 & 0.25091 \\
    \hline
\end{tabularx}

\section{TensorRT Evaluation Numeric Data}
\label{trt_data}

\begin{tabularx}{\linewidth}{|l|X|X|X|X|}
    \hline
     \textbf{Configuration} & \textbf{Avg Runtime (sec)} & \textbf{Stdev Runtime}\\
    \hline
     PyTorch RN50 & 0.2443 & 0.00119 \\
    \hline
     \ic{torch.fx} TensorRT RN50 & 0.0662 & 0.00022 \\
    \hline
     PyTorch LearningToPaint & 0.0068 & 0.0003 \\
    \hline
     \ic{torch.fx} TensorRT LearningToPaint & 0.0044 & 0.0001 \\
    \hline
\end{tabularx}

\end{document}